\documentclass[runningheads]{llncs}
\pdfoutput=1
\usepackage[numbers]{natbib}
\usepackage{multirow,url}
\usepackage{amssymb}
\usepackage{graphicx}
\begin{document}
\mainmatter

\title{Adaptive Spam Detection Inspired by a Cross-Regulation Model of Immune Dynamics: A Study of Concept Drift}
\titlerunning{Adaptive Immuno-Inspired Spam Detection}
\author{Alaa Abi-Haidar$^{*}$ \and Luis M. Rocha}
\authorrunning{Alaa Abi-Haidar and Luis M. Rocha}

\institute{ Department of Informatics, Indiana University, Bloomington IN 47401, USA \\
\and Instituto Gulbenkian de Ci\^{e}ncia, Oeiras, Portugal \\
$^{*}$aabihaid@indiana.edu
}

\maketitle

\begin{abstract}
This paper proposes a novel solution to spam detection inspired by a model of the adaptive immune system known as the cross-regulation model. We report on the testing of a preliminary algorithm on six e-mail corpora. We also compare our results statically and dynamically with those obtained by the Naive Bayes classifier and another binary classification method we developed previously for biomedical text-mining applications. We show that the cross-regulation model is  competitive against those and thus promising as a bio-inspired algorithm for spam detection in particular, and binary classification in general.

\end{abstract}
\section{Introduction}

Spam detection is a binary classification problem in which e-mail is classified as either ham (legitimate e-mail) or spam (illegitimate or fraudulent e-mail). Spam is very dynamic in terms of advertising new products and finding new ways to defeat anti-spam filters. The challenge in spam detection is to find the appropriate threshold between ham and spam leading to the smallest number of misclassifications, especially of legitimate e-mail (false negatives). To avoid confusions, ham and spam will be labeled as negatives and positives respectively.

The vertebrate adaptive immune system, which is one of the most complex and adaptive biological systems, learns to distinguish harmless from harmful substances (known as pathogens) such as viruses and bacteria that intrude the body. These pathogens often evolve new mechanisms to attack the body and its immune system, which in turn adapts and evolves to deal with changes in the repertoire of pathogen attacks. A weakly responsive immune system is vulnerable to attacks while an aggressive one can be harmful to the organism itself, causing autoimmunity. Given the conceptual similarity between the problems of spam and immunity, we investigate the applicability of the cross-regulation model of regulatory T-cell dynamics \cite{carneiro2007tnc} to spam detection.

%
%
Spam detection has recently become an important problem with the ubiquity of e-mail and the rewards of no-cost advertisement that can reach the largest audience possible. Spam detection can target e-mail headers (e.g. sender, receiver, relay servers...) or content (e.g. subject, body). Machine learning techniques such as support vector machines \cite{kolcz2001sbf}, Naive Bayes classifiers \cite{sahami1998baf, metsis2006sfn} and other classification rules such as Case-Based Reasoning \cite{fdezriverola2007sib} have been very successful in detecting spam in the past. However, they generally lack the ability to track \emph{concept drift} since they rely on training on fixed corpora, features, and rules. Concept drift is the (gradual or sudden) change of thematic context (often re-occurring) over time such as new advertisement themes in spam and Bayesian poisening, a technique used by spammers to surpass bayesian based spam filters. Ideally, a system is capable of handling concept drift if it adapts quickly to the thematic change, distinguishing it from noise \cite{tsymbal2004pcd}. Research in spam detection is now focusing on detecting concept drifts in spam, with very promising results \cite{delanycs06, mendez2006tcd}. Other areas of intense development in spam-detection are social-based spam detection models \cite{boykin2005lsn, chirita2005mur} as well as algorithms based on Artificial Immune System (AIS) \cite{oda2005sda} (based on clonal selection) \cite{bezerra2006ifs} (based on ABNET, an AntiBody Network) \cite{yue2007ais} (based on incremental clustering Immune Networks). The AIS models are inspired by diverse responses and theories of the natural immune system \cite{hofmeyr2001iii} such as negative selection, clonal selection, danger theory and the immune network theory. Our bio-inspired spam detection algorithm is based instead on the cross-regulation model \cite{carneiro2007tnc}, which is a novel development in AIS approaches to spam detection. Since this dynamic model is quite compelling in the simplicity by which it achieves harmful/nonharmful\footnote{Or less accurately but more commonly used, self/nonself discrimination} discrimination, we expect it to be useful in also in spam/ham e-mail classification. Moreover, its dynamic nature, in principle, makes it a good candidate algorithm to deal with concept drift in e-mail, which we start testing here.

Section \ref{sec:Introduction} offers a short review of the cross-regulation model \cite{carneiro2007tnc}. Section \ref{sec:ICRM} presents the Cross-regulation Spam Algorithm---our bio-inspired cross-regulation algorithm---and its application to the spam classification problem. Section \ref{sec:Results} discusses the experiments and implementation of the model vis a vis other binary classification models. Finally, in the last two sections, the discussion of the results and the conclusion follow.

\section{The Cross-regulation Model}
\label{sec:Introduction}
The cross-regulation model, proposed by Carneiro et al. \cite{carneiro2007tnc}, aims to model the process of discriminating between harmless and harmful antigens\footnote{Antigens are foreign substances, usually proteins or protein fragments, that trigger immune responses.}---typically harmless self/nonself and harmful nonself. The model consists of only three cell types: Effector T-Cells (E), Regulatory T-Cells (R) and Antigen Presenting Cells (A) whose populations interact dynamically, ultimately to detect harmful antigens. E and R are constantly produced, while A are capable of presenting a collection of antigens to the E and R. T-cell proliferation depends on the co-localization of E and R as they form conjugates (bind) with the antigens presented by A cells (this model assumes that A can form conjugates with a maximum of two E or R). The population dynamics rules of this model are defined by three differential equations, which can be, for every antigen being presented by an A, summarized by the following three laws of interaction:
\begin{enumerate}
\item If one or two E bind to antigen, they proliferate with a fixed rate.
\item If one or two R bind to the antigen, they remain in the population.
\item if an R binds together with an E to the same antigen, the R
 proliferates with a certain rate and the E remains in the population but does not proliferate.
\end{enumerate}

 Finally, the E and R die at a fixed death rate. Carneiro et al. \cite{carneiro2007tnc} showed that the dynamics of this system leads to a bistable system of two possible stable population concentration attractors: (i) the co-existence of both E and R types identifying harmless self antigens, or (ii) the progressive disappearance of R, identifying harmful antigens. 



%
\section{The Cross-regulation Spam Algorithm}
\label{sec:ICRM}
In order to adopt the cross-regulation algorithm for spam detection, which we named the Immune Cross-Regulation Model (ICRM), one has to think of e-mails as analogous to the organic substances that upon entering the body are broken into constituent pieces by lysosome in A. In biology, these pieces are antigens (typically protein fragments) and in our bio-inspired algorithm they are words or features extracted from e-mail messages. Thus, in this model, antigens are words or potentially other features (e.g. bigrams, e-mail titles). For every antigen there exists a number of virtual E and R that interact with A, each associated with a specific e-mail message, and whose role is to present, in distinct slots, a sample of the features of the respective e-mail message. Therefore  A, E and R have specific affinities $\rho \in \Sigma$, where $E_{\rho1}$ and $R_{\rho2}$ can bind to a slot of A, $A_{\rho3}$, only if $\rho1=\rho3$ and $\rho2=\rho3$ respectively. 

The general ICRM algorithm is designed to be first trained on $N$ e-mails of ``self'' (a user's outbox) and harmless ``nonself'' (a user's inbox). However, in the results described here, it was not possible to directly obtain outbox data. We are working on collecting outbox data for future work. Similarly, the ICRM is also trained on ``harmful nonself'' (spam arriving to a given user). Training on or exposure to ham e-mails, in analogy with Carneiro's et al model \cite{carneiro2007tnc}, is supposed to lead to a ``healthy'' dynamics denoted by the co-existence of both E and R with more of the latter. In contrast, training on or exposure to spam e-mails is supposed to result in much higher numbers of E than R. When e-mail features occur for the first time, a fixed initial number of E and R, for every feature, are generated. These initial values of E and R are different in the training and testing stages; more weight to R for ham features, and more weight to E for spam features is given in the labeled training stage. While we specify different values for initializing the proportions of E and R  associated with e-mail features, depending on whether the algorithm is in the training or the testing stage, the ICRM is based on the exact same algorithm in both stages. The ICRM algorithm begins when an e-mail is received and cycles through three phases for every received e-mail: 


\begin{description}
 \item In the \textbf{pre-processing phase}, HTML tags are not stripped off and are treated as other words, as often done in spam-detection \cite{metsis2006sfn} . All words constituting the e-mail subject and body are lowercased and stemmed using Porter's algorithm after filtering out common English stop words and words of length less than 3 characters. A maximum of $n$ processed unique features (words, in this case) are randomly sampled and presented by the virtual A which corresponds to the e-mail. These virtual antigen presenting cells have $n_A$ binding slots (that E and R can bind to) per feature, i.e. $n \times n_A$ slots per e-mail message. The breaking up of the e-mail message into constituent portions (features) is inspired by the natural process in Biology, but is further enhanced in this model to select the first and last $n \over 2$ features in the e-mail. The assumption is that the most indicative information is in the beginning (e.g. subject) and the end of the e-mail (e.g. signature), especially concerning ham e-mails. 

 \item In the \textbf{interaction phase}, feature-specific $R_g$ and $E_f$ are allowed to bind to the corresponding antigens presented by A, which are arbitrarily (uniform random) located on its array of feature slots. Every adjacent pair of A slots is dealt with separately: the $E_f$ for a given feature $f$ proliferate only if they do not find themselves sharing the same adjacent pair of A binding slots with $R_g$, in which case only the $R_g$, associated with feature $g$, proliferate. The model assumes that novel ham features $k$ tend to have their $E_k$ suppressed by $R_g$ of other pre-occurring ham features $g$ because they tend to co-occur in the same message. As for the algorithm's parameters, let $n_A$ be the number of A slots per feature. Let ($E_{0_{ham}}$, $R_{0_{ham}}$) and ($E_{0_{spam}}$, $R_{0_{spam}}$) be the initial values of E and R for features occurring for the first time in the training stage for ham and spam, respectively. For the testing stage, we have ($E_{0_{test}}$, $R_{0_{test}}$). Moreover, $E_{0_{ham}} << R_{0_{ham}}$, $E_{0_{spam}} > R_{0_{spam}}$ and $E_{0_{test}} > R_{0_{test}}$. In the ICRM implementation hereby presented, a major difference form Carneiro's et al model \cite{carneiro2007tnc} was tried: the elimination of cell death. This is a rough attempt to provide the system with long term memory. Cell death can lead to the forgetfulness of spam or ham features if these features do not reoccur in a certain period of time as shown later section \ref{sec:Results}.

 \item In the \textbf{decision phase}, the arriving e-mail is assessed based on the relative proportions of R and E for its $n$ sampled features. Features with more R are assumed to correspond to ham while features with more E are more likely to correspond to spam. The proportions are then normalized to avoid decisions based on a few highly frequent features that could occur in both ham and spam classes. For every feature $f$, the feature score is computed as follows:

 \begin{equation}
 score_{f} = {{R_f - E_f} \over { \sqrt{R_f^2 + E_f^2} }},
  \label{eq:one}
 \end{equation}

 indicating an unhealthy (spam) feature when $score_{f} \le 0$ and a healthy (ham) one otherwise. $score_{f}$ varies between -1 and 1. For every e-mail message $e$, the e-mail immunity score is simply:

\begin{equation}
score_{e} = \sum_{\forall f \in e}{score_{f}}.
 \label{eq:two}
 \end{equation}

 Note that a spam e-mail with no text such as as the cases of messages containing exclusively image and pdf files, which surpass many spam filters, would be classified as spam in this scheme---e-mail $e$ is considered spam if $score_e = 0$. Similarly, e-mails with only a few features occurring for the first time, would share the same destiny, since the initial E is greater than R in the testing stage $E_{0_{test}} > R_{0_{test}}$ which would result in $score_e < 0$.

 \end{description}


%
\section{Results}
\label{sec:Results}

\subsubsection*{E-mail Data}
Given the assumption that personal e-mails (i.e. e-mails sent or received by one specific user) are more representative of a writing style, signature and themes, it would be preferable to test the ICRM on e-mails from a personal mailbox. Unfortunately, this is not offered by the most common spam corpus of \emph{spamassasin}\footnote{http://spamassassin.apache.org/publiccorpus/} and similarly for \emph{ling-spam}\footnote{http://www.aueb.gr/users/ion/publications.html}.  In addition, the ICRM algorithm requires timestamped e-mails, since order of arrival affects final E/R populations. Timestamped data is also important for analyzing concept drifts over time, thus we cannot use the \emph{PU1}\footnote{http://www.iit.demokritos.gr/skel/i-config/downloads/enron-spam/ } data described by Androutsopoulos et al. \cite{androutsopoulos2000ecn} . Delany's spam drift dataset\footnote{http://www.comp.dit.ie/sjdelany/Dataset.htm }, introduced by Delany et al. \cite{DelanyKBS05}, meets the requirements in terms of timestamped and personal ham and spam however its features are hashed and therefore it is not easy to make tangible conclusions based on their semantics. The \emph{enron-spam}\footnote{http://www.iit.demokritos.gr/~ionandr/publications/ } preprocessed data perfectly meets the requirements as it has six personal mailboxes made public after the enron scandal. The ham mailboxes belong to the employees \emph{farmer-d, kaminski-v, kitchen-l, williams-w3, beck-s and lokay-m}. Combinations of five spam datasets were added to the ham data from \emph{spamassassin} (s), \emph{HoneyProject} (h), \emph{Bruce Guenter} (b) and \emph{Georgios Paliousras'} (g) spam corpora and then all six datasets were tokenized \cite{metsis2006sfn}. In practice, some spam e-mails are personalized, which unfortunately cannot be captured in this dataset since the spam data comes from different sources.  Only the first 1500 e-mails of every enron are used in this experiment.


%
\subsubsection*{Evaluation.}
Two forms of evaluation were conducted: The first and more common in spam detection evaluation is the static or offline evaluation using K-fold cross validation \cite{feldman2006tmh} while the second is the dynamic or real-time evaluation using a sliding window that is particularly useful to study the model's capability of dealing with concept drifts in spam and/or ham over time.

In the \textbf{first evaluation}, for each of the six enron sets, we ran each algorithm 10 times. Each run consisted of 200 training (50\% spam) and 200 testing or validation (50\% spam) e-mails that follow in the timestamp order. From the 10 runs we computed variation statistics for the F-score\footnote{The F1-measure (or \emph{F-Score}) is defined as  $F = {{2 \cdot Precision \cdot Recall} \over {Precision+Recall}}$, where $Precision = {TP \over (TP+FP)}$ and $Recall =  {TP \over (TP+FN)}$ and $Accuracy = {(TP+TN) \over (TP+TN+FP+FN)}$ measures of the classification of each test set, where TP, TN, FP and FN denote true positives, true negatives, false positive and false negatives respectively \cite{feldman2006tmh}.}, and Accuracy performance. 

In the \textbf{second evaluation}, for each of the six enron sets, we trained each algorithm on the first 200 e-mails (50\% spam) and then tested on a sliding window of 200 testing or validation (50\% spam) e-mails that follow in the order of time the email was received. The sliding shift was 10 e-mails and the range was between e-mail 201 and e-mail 2800 resulting in 260 slides (from 1500 ham and 1500 spam only 100 ham and 100 spam are for training and the remaining 2800 are for validation). For every window we computed variation statistics of the percentage of FP (misclassified ham) and FN (misclassified spam) in addition to the F-score and Accuracy. We also performed a linear regression of the proportions of false positives and false negatives, \%FP and \%FN, using least squares and computed the slope coefficients, the coefficient of determination $R^2$ for each---for the purpose of evaluating the effect of concept drift if any.

\subsubsection{ICRM Settings.}
In the e-mail pre-processing phase, we used $n=50$, $n_A=10$, $E_{0_{ham}}=6$, $R_{0_{ham}}=12$, $E_{0_{spam}}=6$, $R_{0_{spam}}=5$, $E_{0_{test}}=6$ and $R_{0_{test}}=5$. These initial E and R populations for features occurring for the first time are chosen based on the initial ratios chosen by Carneiro et al. \cite{carneiro2007tnc} and were then empirically adjusted to achieve the best F-score and Accuracy results for the six enron datasets. Finally, the randomization seed was fixed in order to compare results to other algorithms and search for better parameters.

The ICRM was compared with two other algorithms that are explained in the following two subsections. The ICRM was also tested on shuffled (not in order of date received) validation sets to study the importance of e-mail reception order. The results are shown in table \ref{table:two}.

\subsubsection*{Naive Bayes (NB).}
We have chosen to compare our results with the multinomial Naive Bayes with boolean attributes \cite{jensen1996ibn} which has shown great success in previous research \cite{metsis2006sfn}. In order to fairly compare NB with ICRM, we selected the first and last unique $n=50$ features. The Naive Bayes classifies an e-mail as spam in the testing phase if it satisfies the following condition:

\begin{equation}
{{  p(c_{spam}). \prod_{f \in e-mail}{p(f | c_{spam})}} \over  {p(c_{spam}). \sum_{c \in \{c_{spam},{c_{ham}}\}}{ \prod_{f \in e-mail} \\
{p(f | c) }}}} > 0.5,
 \label{eq:three}
\end{equation}

\noindent where $f$ is the feature sampled from an e-mail, and $p(f | c_{spam})$ and $p(f | c_{ham})$ are the probabilities that this feature $f$ is sampled from a spam and ham e-mail respectively, while $c$ is the union of spam and ham e-mails.  The results are shown in table \ref{table:two} and plotted in figure \ref{fig:result}.

\subsubsection*{Variable Trigonometric Threshold (VTT).}
\begin{figure}[ht]
\begin{center}
    \includegraphics[width=10cm,height=10cm]{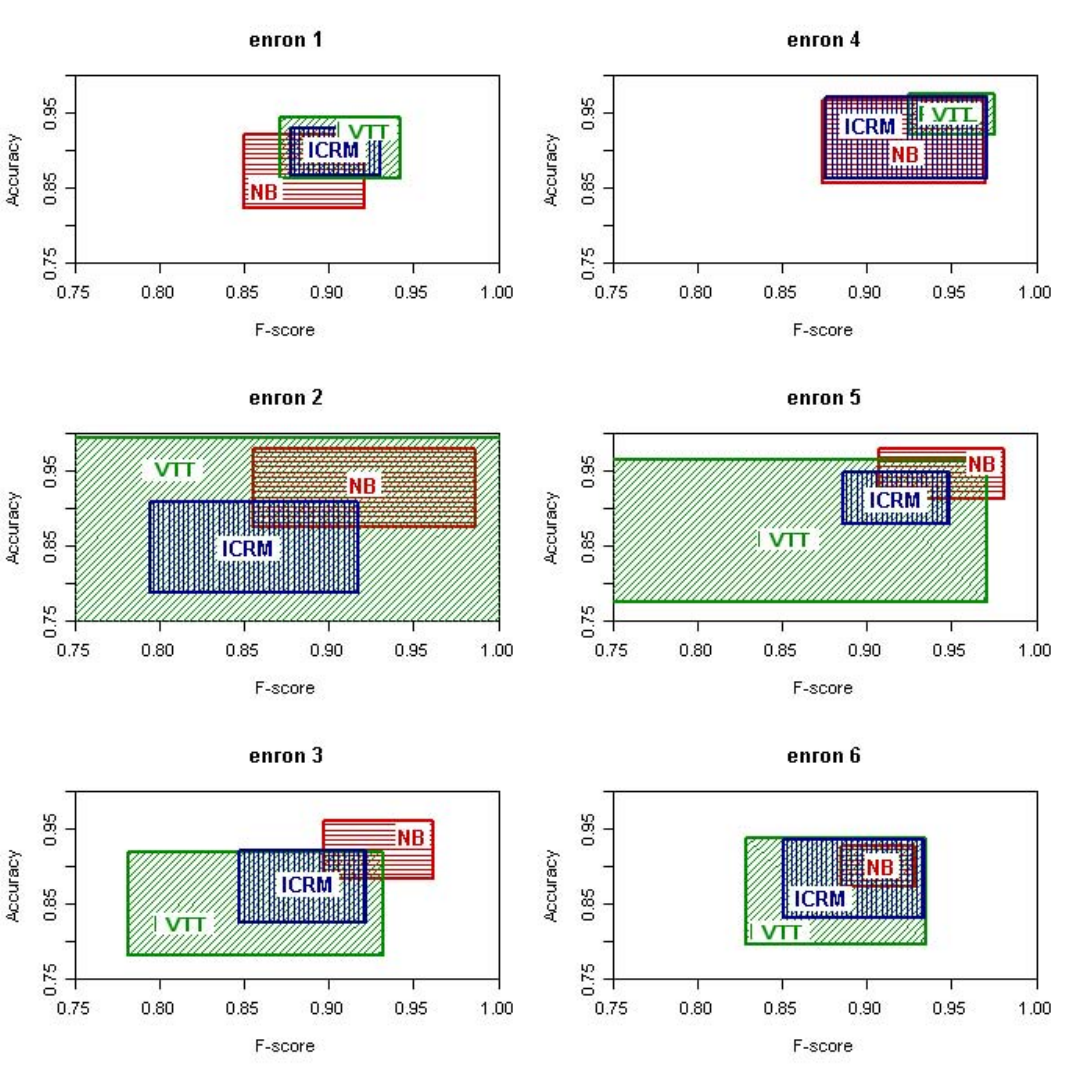}
  \end{center}
vspace{2.5cm}
\caption{\footnotesize F-score vs Accuracy means and standard deviation plot comparison between ICRM (vertical blue), NB (horizontal red) and VTT (diagonal green) for each of the six enron datasets. A visualization of table \ref{table:two}. }
\label{fig:result}
\end{figure}

We previously developed the VTT as a linear binary classification algorithm and implemented it as a protein-protein abstract classification tool\footnote{The Protein Interaction Abstract Relevance Evaluator (PIARE) tool is available at http://casci.informatics.indiana.edu/PIARE/ } using bioliterature mining \cite{abihaidar_GB}.  For more details please refer to \cite{abihaidar_GB}. The results are shown in table \ref{table:two}, plotted in figure \ref{fig:result}.

\begin{table}[h]
\caption{\footnotesize  F-score and Accuracy mean +/- sdev of 10 runs for 50\% spam enron data sets with the first three columns using ICRM (the first one applied on ordered e-mail, the second one on shuffled timestamps of testing data and the third on on ordered e-mail but with ICRM having cell death with death rate=0.02), the fourth one using Naive Bayes and the last one using VTT.}
 \begin{tabular}{|ll||l|l|l|l|l|}
  \hline
 & & \multicolumn{3}{l|}{ICRM} & \multicolumn{2}{l|}{Other Algorithms} \\
\hline
  \textbf{Dataset} & & \textbf{Ordered} & \textbf{Shuffled} & \textbf{Cell Death} &  \textbf{Naive Bayes} & \textbf{VTT} \\
  \hline
  \multirow{2}{*}{Enron1} & F-score & 0.9 $\pm$ 0.03   & 0.9 $\pm$ 0.03 & 0.89 $\pm$ 0.03  & 0.89 $\pm$ 0.04   & 0.91 $\pm$ 0.04    \\
    			  & Accuracy & 0.9 $\pm$ 0.03   & 0.9 $\pm$ 0.03 & 0.89 $\pm$ 0.04     & 0.87 $\pm$ 0.05   & 0.9 $\pm$ 0.04   \\
\hline
  \multirow{2}{*}{Enron2} & F-score & 0.86 $\pm$ 0.06   & 0.85 $\pm$ 0.06 & 0.85 $\pm$ 0.05   & 0.92 $\pm$ 0.07   & 0.82 $\pm$ 0.23  \\
 			  & Accuracy & 0.85 $\pm$ 0.06  & 0.83 $\pm$ 0.07 & 0.84 $\pm$ 0.05    & 0.93 $\pm$ 0.05   & 0.86 $\pm$ 0.13  \\
\hline
  \multirow{2}{*}{Enron3} & F-score & 0.88 $\pm$ 0.04  & 0.88 $\pm$ 0.04 & 0.9 $\pm$ 0.03   & 0.93 $\pm$ 0.03   & 0.86 $\pm$ 0.08  \\
                          & Accuracy & 0.87 $\pm$ 0.05  & 0.87 $\pm$ 0.05 & 0.89 $\pm$ 0.04      & 0.92 $\pm$ 0.04   & 0.85 $\pm$ 0.07  \\
\hline
  \multirow{2}{*}{Enron4} & F-score& 0.92 $\pm$ 0.05   & 0.92 $\pm$ 0.04 & 0.91 $\pm$ 0.06    & 0.92 $\pm$ 0.05   & 0.95 $\pm$ 0.03   \\
   			  & Accuracy& 0.92 $\pm$ 0.05   & 0.92 $\pm$ 0.05  & 0.9 $\pm$ 0.07    & 0.91 $\pm$ 0.06   & 0.95 $\pm$ 0.03  \\
\hline
  \multirow{2}{*}{Enron5} & F-score& 0.92 $\pm$ 0.03    & 0.87 $\pm$ 0.06 & 0.86 $\pm$ 0.04  & 0.94 $\pm$ 0.04   & 0.84 $\pm$ 0.13    \\
   			  & Accuracy& 0.91 $\pm$ 0.03    & 0.87 $\pm$ 0.05  & 0.86 $\pm$ 0.05   & 0.95 $\pm$ 0.03   & 0.87 $\pm$ 0.09    \\
\hline
  \multirow{2}{*}{Enron6} & F-score& 0.89 $\pm$ 0.04     & 0.9 $\pm$ 0.04 & 0.89 $\pm$ 0.03   & 0.91 $\pm$ 0.02   & 0.88 $\pm$ 0.05     \\
   			  & Accuracy& 0.88 $\pm$ 0.05     & 0.89 $\pm$ 0.05  & 0.89 $\pm$ 0.04  & 0.9 $\pm$ 0.03   & 0.87 $\pm$ 0.07    \\
\hline
  \multirow{2}{*}{\textbf{Total}} & \textbf{F-score}& \textbf{0.9 $\pm$ 0.05}     & \textbf{0.89 $\pm$ 0.05} &  \textbf{0.88 $\pm$ 0.05}    &  \textbf{0.92 $\pm$ 0.04}   &  \textbf{0.88 $\pm$ 0.12}     \\
   			         & \textbf{Accuracy}&  \textbf{0.89 $\pm$ 0.05}     &  \textbf{0.88 $\pm$ 0.06} &  \textbf{0.88 $\pm$ 0.05}   &  \textbf{0.91 $\pm$ 0.05}   &  \textbf{0.88 $\pm$ 0.08}    \\
  \hline
\end{tabular}
\label{table:two}
\end{table}

\begin{table}[h]
\caption{\footnotesize  ICRM vs NB F-score and Accuracy for spam to ham ratio variations for all enrons.}
 \begin{tabular}{|ll||l|l|l|}
  \hline
 & & 50\% spam & 30\% spam & 70\% spam \\
  \hline
 \multirow{2}{*}{ICRM}  & F-score &  0.9 $\pm$ 0.05 & 0.91 $\pm$ 0.03  & 0.79 $\pm$ 0.12  \\
                        & Accuracy & 0.89 $\pm$ 0.05 & 0.86 $\pm$ 0.05 & 0.83 $\pm$ 0.08  \\
\hline
 \multirow{2}{*}{NB}    & F-score & 0.92 $\pm$ 0.04 & 0.86 $\pm$ 0.07 & 0.79 $\pm$ 0.07 \\ 
                        & Accuracy & 0.91 $\pm$ 0.05 & 0.84 $\pm$ 0.07  & 0.74 $\pm$ 0.01 \\
  \hline
\end{tabular}
\label{table:three}
\end{table}



\begin{table}[h]
\caption{\footnotesize  ICRM vs NB F-score, accuracy, \%FP and \%FN slope coefficient ($\alpha_{\%FP}$ and $\alpha_{\%FN}$) and $R^2$, \%FP and \%FN for all enrons over time.}
 \begin{tabular}{|ll||l|l|l|l|l|l|}
\hline
  \textbf{Dataset} & & \textbf{F-score} & \textbf{Accuracy} & \textbf{$\alpha_{\%FP}$,$R^2$} &  \textbf{$\alpha_{\%FN}$, $R^2$} &  \textbf{\%FP} & \textbf{\%FN} \\
  \hline
  \multirow{2}{*}{Enron1} & ICRM & 0.95 $\pm$ 0.01   & 0.95 $\pm$ 0.01 & 0.00,0.01  & 0.02,0.41   & 6.7 $\pm$ 1.5 & 4.11 $\pm$ 1.66   \\
    			  & NB & 0.93 $\pm$ 0.01   & 0.93 $\pm$ 0.01 & 0.00,0.27     & 0.03,0.56   & 1.55 $\pm$ 0.53 &  12.99 $\pm$ 2.7 \\
\hline
  \multirow{2}{*}{Enron2} & ICRM & 0.92 $\pm$ 0.01    &0.92 $\pm$ 0.01  & 0.00,0.01   & -0.01,0.11   & 6.48 $\pm$ 1.17  & 8.87 $\pm$ 1.89 \\
 			  & NB & 0.95 $\pm$ 0.01  & 0.94 $\pm$ 0.01 & 0.01,0.10   &0.00,0.01   &  9.57 $\pm$ 2.05  & 1.29 $\pm$ 0.48 \\
\hline
  \multirow{2}{*}{Enron3} & ICRM & 0.93 $\pm$ 0.02   & 0.94 $\pm$ 0.02  & 0.03,0.95   & 0.01,0.20   & 4.7 $\pm$ 2.06 &  8.37 $\pm$ 1.77 \\
                          & NB & 0.92 $\pm$ 0.03  & 0.92 $\pm$ 0.02 &   0.00,0.43     & 0.05,0.52   &  0.51 $\pm$ 0.4 & 16.2 $\pm$ 5.2 \\
\hline
  \multirow{2}{*}{Enron4} & ICRM & 0.92 $\pm$ 0.03   & 0.92 $\pm$ 0.03 & 0.04,0.52    &0.03,0.37  & 6.99 $\pm$ 4.03 &  9.99 $\pm$ 2.92 \\
   			  & NB & 0.92 $\pm$ 0.01   & \textbf{0.93 $\pm$ 0.01}  & 0.00,0.56, & 0.04,0.63 & 0.18 $\pm$ 0.27&  15 $\pm$ 3.06 \\
\hline
  \multirow{2}{*}{Enron5} & ICRM& 0.90 $\pm$ 0.02    & 0.90 $\pm$ 0.02 & 0.03,0.49  & 0.02,0.49   &  8.54 $\pm$ 2.58 & 12.08 $\pm$ 2.1    \\
   			  & NB& 0.96 $\pm$ 0.03    & 0.96 $\pm$ 0.03  & 0.02,0.22  & 0.04,0.77    & 4.76 $\pm$ 3.44 &  3.06 $\pm$ 3.1   \\
\hline
  \multirow{2}{*}{Enron6} & ICRM & 0.93 $\pm$ 0.01     & \textbf{0.93 $\pm$ 0.02} & 0.03,0.85   & 0.01,0.28   & 8.09 $\pm$ 2.23 & 5.33 $\pm$ 1.23  \\
   			  & NB & 0.95 $\pm$ 0.01     & 0.95 $\pm$ 0.01  & 0.01,0.06 & 0.00,0.09  &  3.07 $\pm$ 2.17 &  6.89 $\pm$ 1.04   \\
\hline
\end{tabular}
\label{table:four}
\end{table}

\begin{figure}[h]
\begin{center}
    \includegraphics[width=12cm,height=6cm]{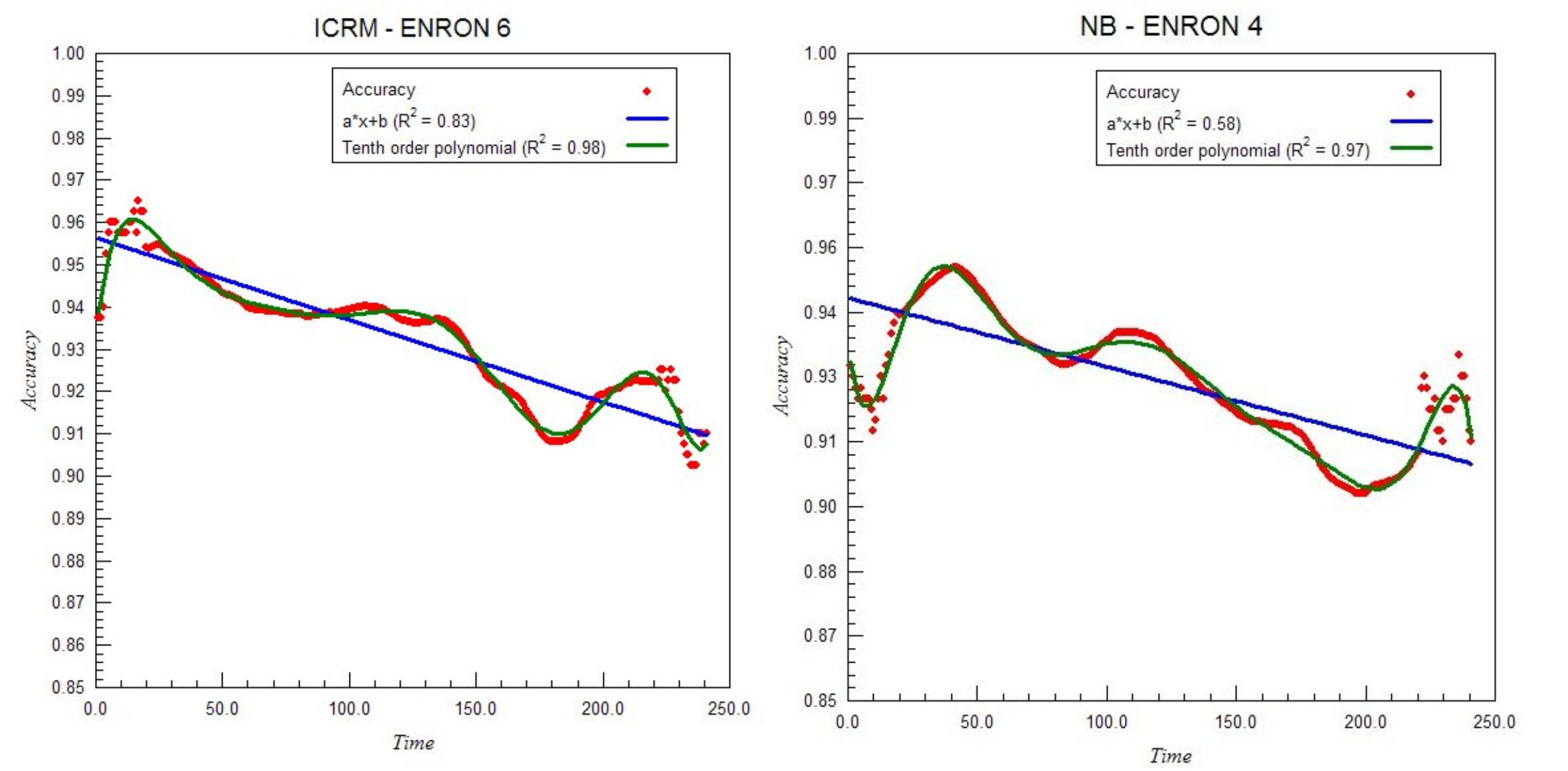}
  \end{center}
\caption{\footnotesize ICRM Accuracy over time for enron6 and NB Accuracy over time for enron4, showing best linear and polynomial fits with $R^2$. The rest of the Accuracy and FN/FP plots are available as supplementary material.}
\label{fig:enron6}
\end{figure}

\section{Discussion}
\label{sec:Discussion}

\subsubsection*{Static Evaluation Results.}

As clearly shown in table \ref{table:two}, ICRM, NB and VTT are very competitive for most enron datasets, indeed the performance of ICRM is statistically indistinguishable from VTT (F-score and Accuracy p-values 0.15 and 0.63 for the paired t-test validating the null hypothesis of variation equivalence), though its slightly lower performance against NB is statistically significant (F-score and Accuracy p-values 0.01 and 0.02 for the paired t-test, rejecting the null hypothesis of variation equivalence with 0.05 level of significance).
 
However, the ICRM can be more resilient to ham ratio variations\footnote{ The 30\% and 70\% spam results were balanced for the evaluation by randomly sampling from the 70\% class, reducing it to 30\%.} as shown in table \ref{table:three} and figure \ref{fig:spamratio}. While the performance of both algorithms was comparable for 50\% spam (though significantly better for NB), the performance of NB drops for 30\% spam ratio (5\% lower F-score than ICRM) and 70\% spam ratio (9\% less accurate than ICRM) while ICRM relatively maintains a good performance. The difference in performance is statistically significant, except for F-Score of the 70\% spam experiment, as the p-values obtained for our performance measures clearly reject the null hypothesis of variation equivalence: F-Score and Accuracy p-values are 0 and 0.01 for 30\% spam, and Accuracy p-value is 0.01 for 70\% spam (p-value for F-Score is 0.5 for this case). While one could argue that NB's performance could well be increased, in the unbalanced spam/ham ratio experiments, by changing the right hand side of equation \ref{eq:three} to 0.3 or 0.7, this act would imply that, in real situations, one could know a priori the spam to ham ratio of a given user. The ICRM model, on the other hand, does not need to adjust any parameter for different spam ratios---it is automatically more reactive to whatever ratio it encounters. It has been shown that spam to ham ratios indeed vary widely \cite{meyer2004seo, DelanyKBS05}, hence we conclude that the ICRM's ability to better handle unknown spam to ham ratio variations is more preferable for dynamic data classification in general and spam detection in particular.  

We have implemented ICRM with death rate\footnote{Death rate = 0.02 resulted in the best performance for the death rate range $r \in [0.01, 0.1]$, where $r$ is the probability that an $R_f$ or $E_f$ would die for a previously occurring feature $f$.} = 0.02. and without virtual cell death but the results showed negligible statistical differences (F-score and Accuracy p-values 0.02 and 0.04) although slightly in favor of no virtual cell death, as seen in table \ref{table:two}. The ICRM tested for spam variation and dynamic evaluation excluded cell death to speed up the algorithm, nonetheless, we are in the process of experimenting with heterogeneous death rates for the E, R cells of different features and more ``interesting'' features (e.g. e-mail title, from, to, and cc features). Since death rates affect the long-term memory of the system, this is something we intend to investigate more closely in future work.  

In most Enron sets, shuffling the timestamps of received e-mails in the testing stages also only slightly reduced the ICRM's performance (F-score and Accuracy p-values 0.07 and 0.04 for paired t-test), therefore, unlike what was expected, the timestamps of e-mails seem to be largely irrelevant---which undermines some of the justification for a dynamic approach to spam detection based on the cross-regulation model. Nevertheless, we plan to study this further with additional data sets with much longer date ranges. 

\subsubsection*{Dynamic Evaluation Results.}

The ICRM was also very competative with NB, have shown to be very competitive in the dynamic evaluation. The evidence is in the first two columns (F-score and Accuracy) of table \ref{table:four} and in the supplementary material section\footnote{All supplementary material is accessible at http://casci.informatics.indiana.edu/icaris08/}.

Another notable feature of the ICRM is that it seems to balance \%FN and \%FP more efficiently over time. Conversely, NB tends to have high \%FN and low \%FP or vice versa. In order to quantify the balance between \%FP and \%FN, we compute their means and standard deviations for all enrons in the last two columns of table \ref{table:four}. While the largest mean in ICRM does not exceed 12.08\% (enron 5), it does reach 12.99\% (enron 1)  16.02\% (enron 3) and 15\% (enron 4) in NB for \%FP. However, in spam detection in particular, more importance is given to \%FP (ham misclassification) which favors NB over ICRM in most enron datasets. In future work, we will explore if enabling heterogeneous death rates for E and R cells can reduce \%FP with the ICRM. On the other hand, the ICRM's more balanced \%FN and \%FP could be valuable for other binary classification problems where FP and FN are equally important---which is not the case in spam detection.   

We also computed slope coefficients $\alpha_{\%FN}$, $\alpha_{\%FP}$ and their corresponding $R^2$ for the least square linear fit of \%FN and \%FP in order to study the behaviour of concept drift which would be manifested by high slopes---indicating decay in performance. However,  the slopes are quite minimal as shown in the third and fourth columns of table \ref{table:four}. Indeed, the performance is essentially flat in time for both algorithms with slopes close to zero (see plots in supplemental materials). Therefore, there does not seem to be much concept drift in these datasets.

The observations made based on the artificial immune system can help us guide or further deepen our understanding of the natural immune system. For instance, ICRM's resilience to spam to ham ratio and its ability to balance between \%FN and \%FP show us how dynamic is our immune system and functional independently of the amount of pathogens attacking it. In addition, the three modifications made to the original model can be very insightful: The improvements made by training on both spam and ham (rather than only ham or self) reinforce the theories of both self and nonself antigen recognition by T-cells outside the thymus. The feature selection makes us wonder whether the actual T-cell to antigen binding is absolutely arbitrary. Finally, the elimination of cell death may reinforce the theories behind long lived cells as far as long term memory is concerned.

\section{Conclusion}
\label{sec:Conclusion}
In this paper we have introduced a novel spam detection algorithm inspired by the cross-regulation model of the adaptive immune system. Our model has proved itself competitive with both spam binary classifiers and resilient to spam to ham ratio variations in particular. The overall results, even though not stellar, seem quite promising especially in the areas of spam to ham ratio variation and also of tracking concept drifts in spam detection.  This original work should be regarded not only as a promising bio-inspired method that can be further developed and even integrated with other methods but also as a model that could help us better understand the behavior of the T-cell cross-regulation systems in particular, and the vertebrate natural immune system in general.

\subsubsection*{Acknowledgements.}
We thank Jorge Carneiro for his insights about applying ICRM on spam detection and his generous support and contribution for making this work possible. We also thank Florentino Fdez-Riverola for the very useful indications about spam datasets and work in the area of spam detection. We would also like to thank the FLAD Computational Biology Collaboratorium at the Gulbenkian Institute in Oeiras, Portugal, for hosting and providing facilities used to conduct part of this research.

\scriptsize
%
%

%
\end{document}